\documentclass{article}

\PassOptionsToPackage{numbers,compress}{natbib}
\usepackage[dblblindworkshop,final]{{neurips_2026}} \workshoptitle{{SLMs for Agentic Systems, Paris, France, 2026}}
\usepackage[utf8]{inputenc}
\usepackage[T1]{fontenc}
\usepackage{hyperref}
\usepackage{url}
\usepackage{booktabs}
\usepackage{amsmath,amssymb,amsthm}
\usepackage{nicefrac}
\usepackage{microtype}
\usepackage{xcolor}
\usepackage{colortbl}
\definecolor{modelrow}{gray}{0.90}
\newcommand{\modelheader}[2]{\rowcolor{modelrow}\multicolumn{#1}{l}{\emph{#2}}}
\usepackage{graphicx}

\newcommand{\method}{\textsc{STAC}}
\newcommand{\keep}{\textsc{Keep}}
\newcommand{\switch}{\textsc{Switch}}
\newcommand{\qkeep}{\mathrm{KEEP}}
\newcommand{\qswitch}{\mathrm{SWITCH}}
\newcommand{\ind}{\mathbb{I}}

\title{Hierarchical Reinforcement Learning with Stable Temporal Abstraction for Language Model Agents}
\author{%
  Shayan Mohajer Hamidi\thanks{Corresponding author.}\\
  LinkedIn Corporation\\
  smohajer@linkedin.com \\
  \And
Yize Cheng\thanks{Work done while an intern at LinkedIn Corporation.} \\
  University of Maryland \\
  yzcheng@umd.edu \\
  \And
  Yuanda Xu \\
  LinkedIn Corporation \\
  ericxu@linkedin.com \\
  \AND
  Zhengze Zhou \\
  LinkedIn Corporation \\
  zzhou1@linkedin.com \\
  \And
Alborz Geramifard \\
  LinkedIn Corporation \\
  agf@linkedin.com \\
}

\begin{document}
\maketitle

\begin{abstract}
Hierarchical reinforcement learning improves long-horizon control by organizing primitive actions around persistent subgoals and assigning credit at multiple temporal scales.
Recent hierarchical language agents bring these benefits to interactive tasks by explicitly separating subgoal planning from action execution.
We observe, however, that an explicit hierarchy does not by itself determine how stable the resulting temporal abstraction is: the learned boundary policy may replace the subgoal almost every turn, making it effectively transient, or retain a subgoal after it has stopped being appropriate.
We call this \emph{temporal abstraction instability}.
We propose \emph{Stable Temporal Abstraction via Constrained Optimization} (\method{}), a constrained boundary-policy optimization method that represents premature replanning and stale persistence as constraint costs.
\method{} applies the resulting Lagrangian costs only to the sampled boundary decision, leaving the underlying algorithm's rewards, critic targets, subgoal advantages, and primitive-action advantages unchanged.
Across two backbones and two benchmarks, \method{} improves  success over a strong hierarchical baseline by $8.1$ and $7.9$ points on ALFWorld and WebShop with Qwen3-0.6B, and by $23.5$ and $15.8$ points with Llama-3.2-1B-Instruct.
\end{abstract}

\section{Introduction}
\label{sec:intro}

Hierarchical reinforcement learning (HRL) has long provided a principled approach to long-horizon decision making~\cite{sutton1999options,dietterich2000maxq,nachum2018hiro,li2019haar,zhang2020hrac,gurtler2021hits,lee2022dhrl,corcoll2022cehrl}.
Rather than selecting every primitive action at a single temporal scale, HRL introduces temporally extended subgoals or options and decomposes their realization into lower-level actions~\cite{bacon2017option}.
This decomposition can shorten effective planning horizons, reuse behavior across states, preserve intermediate intent, and separate credit for high-level planning from credit for low-level execution.

Recent work has begun transferring these benefits to language-model agents~\cite{zhou2024archer,hu2025glider,zhen2026stephrl,jin2026himac,diao2026hipif,park2026multi2,peng2026hiper}.
For example, HiPER~\cite{peng2026hiper} represents each interaction turn through an explicit Plan--Execute hierarchy: a high-level component produces a textual subgoal, a boundary decision determines whether that subgoal should persist or be replaced, and a low-level component selects an environment action.
Its hierarchical advantage estimator further assigns distinct learning signals to subgoal, boundary, and action tokens.
Such methods demonstrate that explicit hierarchical structure and multi-timescale credit assignment can improve long-horizon language-agent performance.

We observe, however, that the architecture alone does not determine how stable the learned temporal abstraction will be.
Useful temporal abstraction requires a subgoal to persist long enough to coordinate multiple actions, but also to terminate once it is completed or no longer appropriate.
The learned boundary policy can drift away from either requirement.
It may replace the subgoal almost every turn, reducing the subgoal toward a transient one-step label and moving the hierarchy toward a flat policy.
Alternatively, it may retain the same subgoal too long, delaying state-dependent replanning after that subgoal stops being appropriate.
We call this tendency \underline{\emph{temporal abstraction instability}}.
It matters most in multi-phase tasks, where success requires both persistence within a phase and a well-timed transition between phases.

To address this, we propose \emph{Stable Temporal Abstraction via Constrained Optimization} (\method{}), a constrained boundary-policy optimization method.
\method{} represents premature replanning and stale-subgoal persistence as constraint costs and optimizes their Lagrangian relaxation.
It consumes the normalized boundary advantage supplied by the hierarchical optimizer it is applied to, which we call the \emph{host}, and modifies only the coefficient of the sampled boundary decision.
Environment rewards, critic targets, subgoal advantages, and primitive-action advantages remain unchanged.
This decision locality distinguishes \method{} from reward shaping and makes it independent of the host's particular actor architecture or return estimator.

We evaluate \method{} using the publicly released HiPER Plan--Execute architecture~\cite{peng2026hiper} as a strong hierarchical host, whose boundary policy replans early on more than half of all turns.
\method{} instead maintains multi-turn subgoals and nearly perfect structured-control validity.
The improved hierarchy is accompanied by an $8.1$-point ALFWorld gain over the released host recipe on Qwen3-0.6B, rising to $23.5$ points on Llama-3.2-1B-Instruct.
Our contributions are:

\noindent $\bullet$ We identify and characterize \emph{temporal abstraction instability}: a hierarchy can be architecturally explicit yet still learn subgoals that are too transient or too persistent for useful temporal structure.

\noindent $\bullet$     We formulate stable temporal abstraction as a constrained boundary-policy problem and derive a decision-local Lagrangian optimization rule with separate costs for premature replanning and stale-subgoal persistence; our reported experiments optimize its fixed-multiplier relaxation, and we report the violation rates it induces.

\noindent $\bullet$     We show that constraining the boundary yields longer, valid multi-turn subgoals relative to a released hierarchical host, and that this accompanies improved ALFWorld and WebShop performance on two model backbones.

\section{Hierarchical Setup and Temporal Abstraction Instability}
\label{sec:setup}

We follow the hierarchical formulation of HiPER~\cite{peng2026hiper}, whose Plan--Execute agent and hierarchical advantage estimator we adopt unchanged and use as the host optimizer throughout; this section restates that setup in the notation used later.

\paragraph{Task, action, and reward.}
Let $x\sim p(\mathcal X)$ be an interactive task and let $t\in\{0,\ldots,T-1\}$ index the environment turns in an episode.
At turn $t$, $s_t$ contains the current observation and interaction history available to the policy.
The policy selects a primitive action $a_t$, meaning one environment interaction such as a tool call together with its arguments, and the environment transitions to the next observation.
The environment returns a task reward after the action, and the host optimizer may add fixed shaping terms; $r_t$ denotes the resulting training reward.
The return is
$G_t=\sum_{u=t}^{T-1}\gamma^{u-t}r_u$.
The task objective is
\begin{equation}
J_{\mathrm{task}}(\theta)
=
\mathbb E_{x\sim p(\mathcal X),\,\tau\sim\pi_\theta(\cdot\mid x)}
\left[\sum_{t=0}^{T-1}\gamma^t r_t\right],
\label{eq:task-objective}
\end{equation}
where $\tau$ is the complete interaction trajectory and $\gamma\in(0,1]$ is the discount factor.

\paragraph{Subgoals and temporal segments.}
A subgoal $z_t$ is a textual intermediate objective whose completion normally requires several primitive actions.
For a tool-using language agent, this means several tool calls with concrete arguments.
For example, completing ``heat the object'' may require navigating to an appliance, opening it, placing the object, and activating it.
Consecutive turns governed by the same subgoal form a temporal segment.
The subgoal should persist while its actions are executed, then be replaced when the intermediate objective is complete or no longer appropriate.

\paragraph{Boundary-explicit policy.}
Let $z_{t-1}$ be the subgoal active before turn $t$.
The policy first samples a boundary decision $q_t\in\{\qkeep,\qswitch\}$.
If $q_t=\qkeep$, it deterministically retains $z_t=z_{t-1}$; if $q_t=\qswitch$, it generates a new subgoal $z_t$.
Let $\theta$ denote the trainable actor parameters and let $\pi_\theta$ denote the joint hierarchical policy.
Its factors are the boundary policy $\pi_\theta^b$, the subgoal policy $\pi_\theta^z$, and the primitive-action policy $\pi_\theta^a$:
\begin{align}
\pi_\theta(q_t,z_t,a_t\mid s_t,z_{t-1})   =
\pi_\theta^b(q_t\mid s_t,z_{t-1})  \cdot\pi_\theta^z(z_t\mid s_t)^{\ind[q_t=\qswitch]}
\cdot\pi_\theta^a(a_t\mid s_t,z_t).
\label{eq:factorization}
\end{align}
The indicator exponent removes the subgoal-generation factor on \keep{} turns, when $z_t$ is copied deterministically from $z_{t-1}$.
The factorization may be implemented by separate modules or by one autoregressive model with structured boundary, subgoal, and action spans.

\paragraph{HiPER host objective and advantages.}
Our experiments instantiate Eq.~\eqref{eq:factorization} with HiPER, whose single autoregressive actor emits the boundary, subgoal, and action spans and is trained with PPO.
The policy-gradient form underlying its clipped actor objective separates the three decisions:
\begin{align}
\nabla_\theta J_{\mathrm{task}}(\theta)
\approx \mathbb E\!\left[\sum_{t=0}^{T-1}
\left(
\begin{aligned}
&\nabla_\theta\log\pi_\theta^b(q_t\mid s_t,z_{t-1})\widehat A_t^b\\
&+\ind[q_t=\qswitch]\nabla_\theta\log\pi_\theta^z(z_t\mid s_t)\widehat A_t^z+\nabla_\theta\log\pi_\theta^a(a_t\mid s_t,z_t)\widehat A_t^a
\end{aligned}
\right)\right].
\label{eq:host-gradient}
\end{align}
Here $\widehat A_t^b$, $\widehat A_t^z$, and $\widehat A_t^a$ assign credit to the sampled boundary decision, newly generated subgoal, and primitive action, respectively.
HiPER estimates all three using one shared critic backbone with only two output heads:
with critic parameters $\phi$,
$V_\phi^{H}(s_t)\approx\mathbb E[G_t\mid s_t,q_t=\qswitch]$ for planning at a segment boundary and
$V_\phi^{L}(s_t,z_t)\approx\mathbb E[G_t\mid s_t,z_t]$ for execution under the active subgoal.
The subgoal advantage uses segment-level GAE with $V_\phi^H$; the action advantage uses within-segment GAE with $V_\phi^L$ and bootstraps into $V_\phi^H$ at the next segment boundary.
There is no third boundary-value head.
Instead, with $y_t=\ind[q_t=\qswitch]$ and
$\beta_t=\pi_\theta^b(\qswitch\mid s_t,z_{t-1})$, HiPER estimates
\begin{equation}
\widehat A_t^b
=
(y_t-\beta_t)
\left(V_\phi^H(s_t)-V_\phi^L(s_t,z_{t-1})\right).
\label{eq:host-boundary-advantage}
\end{equation}
Thus the boundary signal compares the estimated return from replanning with that from continuing the previous subgoal, then centers this gain for the sampled binary decision.
Before the PPO update, HiPER standardizes each advantage stream independently. For the boundary stream this yields
\begin{equation}
\overline A_t^b
=
\frac{\widehat A_t^b-\operatorname{mean}_{u\in\mathcal B}\widehat A_u^b}
{\operatorname{std}_{u\in\mathcal B}\widehat A_u^b+\varepsilon},
\label{eq:normalized-boundary-advantage}
\end{equation}
where $\mathcal B$ is the set of turns in the update batch whose boundary-token mask is valid, and $\varepsilon>0$ prevents division by zero.
We refer to $\overline A_t^b$ as the \emph{normalized boundary advantage}.
It is a single scalar per turn, attached to the boundary tokens that realize the $\qkeep$/$\qswitch$ decision, and it is positive when the sampled decision looked better than the boundary policy's own average for that batch.
Two properties matter for what follows.
First, $\overline A_t^b$ is the only host quantity \method{} reads or modifies: the subgoal advantage $\widehat A_t^z$, the action advantage $\widehat A_t^a$, and both critic targets are left untouched, which is what makes the method decision-local.
Second, because standardization is applied per batch, $\overline A_t^b$ carries no information about how long the current subgoal has already been active; it scores the decision only by estimated return.
That omission is precisely the gap \method{} fills, by adding duration-dependent costs to this term.
Appendix~\ref{app:host-advantage} gives the complete estimator and critic targets.

\paragraph{Behavioral criterion and instability.}
Let $d_t\ge1$ count the subgoal-creating turn and all subsequent turns since the most recent valid switch.
Useful temporal abstraction requires state-dependent, non-degenerate duration: the subgoal should span multiple actions when coordination is needed and terminate when it becomes stale.
As $\pi^b$ moves toward always switching, $d_t$ approaches $1$, segment-level returns approach per-turn returns, and the subgoal approaches a transient scratchpad.
As it moves toward always keeping, the agent replans less often than the task requires.
Both directions reduce the temporal structure the hierarchy is meant to provide.

\section{STAC: Stable Temporal Abstraction via Constrained Optimization}
\label{sec:method}

\subsection{Constrained temporal abstraction}

Let $g_{\mathrm{short}}(s_t,z_{t-1},d_t)$ and
$g_{\mathrm{stale}}(s_t,z_{t-1},d_t)$ be bounded cost functions in $[0,1]$ that measure, respectively, whether replanning at turn $t$ would be premature and whether retaining the current subgoal would be stale.
These costs may be fixed, state-dependent, or learned; the general method does not require a universal duration interval.
For a parsed boundary decision, define
\begin{align}
c_t^{\mathrm{short}}
&=\ind[q_t=\qswitch]\,
g_{\mathrm{short}}(s_t,z_{t-1},d_t), \notag\\
c_t^{\mathrm{stale}}
&=\ind[q_t=\qkeep]\,
g_{\mathrm{stale}}(s_t,z_{t-1},d_t).
\label{eq:violation}
\end{align}
At policy update $k$, let $\pi_k=\pi_{\theta_k}$ be the behavior policy and let $\mathcal D_k$ be its on-policy batch.
The boundary likelihood ratio is
$r_t^b(\theta)=\pi_\theta^b(q_t\mid s_t,z_{t-1})/\pi_k^b(q_t\mid s_t,z_{t-1})$.
Define the decision-local task and constraint surrogates
\begin{align}
\mathcal J_k^{\mathrm{task}}(\theta)
&=\mathbb E_{t\sim\mathcal D_k}
\left[r_t^b(\theta)\overline A_t^b\right], \notag\\
\mathcal C_i^k(\theta)
&=\mathbb E_{t\sim\mathcal D_k}
\left[r_t^b(\theta)c_t^i\right],
\quad i\in\{\mathrm{short},\mathrm{stale}\}.
\label{eq:surrogates}
\end{align}
Here $\overline A_t^b$ is the normalized boundary advantage of Eq.~\eqref{eq:normalized-boundary-advantage}, so $\mathcal J_k^{\mathrm{task}}$ is the host's own boundary objective and \method{} adds no new return signal; for a different host, $\overline A_t^b$ may be produced by any boundary-level estimator. At $\theta=\theta_k$, these constraint surrogates equal the observed violation rates in the current batch.
We formulate each policy update as
\begin{align}
\max_\theta\quad &\mathcal J_k^{\mathrm{task}}(\theta) \notag\\
\text{s.t.}\quad
&\mathcal C_{\mathrm{short}}^k(\theta)\le\rho_{\mathrm{short}}, \notag\\
&\mathcal C_{\mathrm{stale}}^k(\theta)\le\rho_{\mathrm{stale}},
\label{eq:constrained}
\end{align}
where the $\rho_i$ specify tolerated violation rates.
Because these surrogates are measurable, the violation rates realized by the learned policy can be read off during training; Fig.~\ref{fig:constraints} reports them for our runs.
Its Lagrangian is
\begin{align}
\mathcal L_k(\theta,\boldsymbol\mu)
=\mathcal J_k^{\mathrm{task}}(\theta) -\sum_{i\in\{\mathrm{short},\mathrm{stale}\}}
\mu_i\bigl(\mathcal C_i^k(\theta)-\rho_i\bigr),
\quad \mu_i\ge0.
\label{eq:lagrangian}
\end{align}
The multipliers in Eq.~\eqref{eq:lagrangian} can be updated by projected dual steps,
\begin{equation}
\mu_i\leftarrow
\left[\mu_i+\eta_\mu\bigl(\widehat{\mathcal C}_i^k-\rho_i\bigr)\right]_+,
\label{eq:dualupdate}
\end{equation}
where $\widehat{\mathcal C}_i^k$ is the batch estimate of the corresponding constraint surrogate.
In the experiments reported here we use a \emph{fixed-multiplier} instantiation of this Lagrangian relaxation, holding $\mu_{\mathrm{short}}=\mu_{\mathrm{stale}}=0.1$ throughout training rather than applying Eq.~\eqref{eq:dualupdate}.
For fixed $\boldsymbol\mu$ the terms $\sum_i\mu_i\rho_i$ are constant in $\theta$, so $\arg\max_\theta\mathcal L_k(\theta,\boldsymbol\mu)=\arg\max_\theta[\mathcal J_k^{\mathrm{task}}(\theta)-\sum_i\mu_i\mathcal C_i^k(\theta)]$ and the budgets $\rho_i$ do not enter the update.
We therefore optimize the primal policy under fixed temporal-abstraction prices; Eq.~\eqref{eq:dualupdate} gives the corresponding adaptive primal--dual extension, which we leave to future work.
For valid boundary decisions, collecting the task and constraint terms yields the sampled advantage
\begin{equation}
\widehat A_t^{b,\mathrm{STAC}}
=
\overline A_t^b
-\mu_{\mathrm{short}}c_t^{\mathrm{short}}
-\mu_{\mathrm{stale}}c_t^{\mathrm{stale}}.
\label{eq:auxadv}
\end{equation}
An on-policy host then uses this combined advantage in its ordinary actor objective; for PPO, it is inserted into the standard clipped surrogate.
Only tokens in the boundary-decision span receive the short- and stale-subgoal terms.
Subgoal and action advantages, task rewards, and critic targets are unchanged.
Appendix~\ref{app:optimization-details} gives the locality analysis, training algorithm, and diagnostics.

Note that \method{} does not add its temporal-abstraction costs to $r_t$; the exact reward choices in our experiments are given in Appendix~\ref{app:reproducibility}.
The reported instantiation is therefore a fixed two-sided boundary penalty, not adaptive dual optimization with user-specified budgets.
We prespecify $d_{\min}=2$ and $d_{\max}=8$ as duration heuristics: the former discourages a uniform one-turn collapse, while the latter is horizon-informed but not a per-segment upper bound (Appendix~\ref{app:thresholds}).
\section{Experiments}
\label{sec:experiments}

\paragraph{Environments and model.}
We evaluate on two text-based interactive benchmarks with sparse success rewards.
ALFWorld~\cite{shridhar2021alfworld} contains long-horizon household tasks whose multi-phase categories require several coordinated actions per objective, and WebShop~\cite{yao2022webshop} requires navigating a simulated e-commerce site to satisfy a natural-language shopping instruction.
Both reward persistent subgoals while still requiring well-timed replanning.
The actor and critic use Qwen3-0.6B~\cite{qwen3} unless stated otherwise; to test whether the effect depends on a single pretraining lineage, we repeat both benchmarks with Llama-3.2-1B-Instruct~\cite{llama32} under identical hyperparameters and seeds.
We instantiate \method{} on the public HiPER Plan--Execute implementation~\cite{peng2026hiper}, whose structured actor emits a boundary decision, textual subgoal, and admissible action, and whose estimator supplies $\overline A^b$; \method{} leaves the subgoal/action streams and all critic targets unchanged (Appendix~\ref{app:host-advantage}).

\paragraph{Training.}
Each run uses 128 independent environments per update, 150 updates, at most 50 environment turns, training temperature $1.0$, and validation temperature $0.4$; actor and critic learning rates are $10^{-6}$ and $10^{-5}$ and the KL coefficient is $0.01$.
We evaluate 128 held-out episodes every five updates, and the base checkpoint before training.
Appendix~\ref{app:reproducibility} gives the full configuration and checkpoint protocols.

\paragraph{Benchmarks.}
We compare against two flat baselines that optimize a single-timescale policy: GRPO~\cite{shao2024grpo}, which normalizes trajectory-level rewards within a sampled group, and GiGPO~\cite{feng2025gigpo}, which adds step-level relative advantages for denser credit.
Our hierarchical host is HiPER~\cite{peng2026hiper}, whose Plan--Execute actor and hierarchical advantage estimator provide the boundary advantage that \method{} modifies.

\section{Results}
\label{sec:results}

\begin{table*}[t]
\centering
\caption{\textbf{Main results across environments and model families.} Values are mean success (\%) $\pm$ SD across three seeds. ``Best'' selects each trained method's highest-validation checkpoint following the released HiPER convention; ``Final'' uses update 150.}
\label{tab:results}
\small
\setlength{\tabcolsep}{10pt}
\begin{tabular}{llrrrr}
\toprule
& & \multicolumn{2}{c}{ALFWorld} & \multicolumn{2}{c}{WebShop} \\
\cmidrule(lr){3-4}\cmidrule(lr){5-6}
Type & Method & Best & Final & Best & Final \\
\midrule
\modelheader{6}{Qwen3-0.6B} \\
--- & Base model
& \multicolumn{2}{c}{$0.3\pm0.5$}
& \multicolumn{2}{c}{$1.2\pm1.7$} \\
Flat & GRPO & $37.5\pm2.8$ & $35.9\pm4.7$ & $44.7\pm3.2$ & $44.7\pm3.2$ \\
Flat & GiGPO & $59.1\pm7.1$ & $59.1\pm7.1$ & $51.3\pm5.5$ & $51.3\pm5.4$ \\
Hier. & HiPER & $83.3\pm3.3$ & $81.5\pm3.6$ & $61.4\pm3.5$ & $56.5\pm3.0$ \\
Hier. & \method{} (ours) & $\mathbf{91.4\pm3.6}$ & $\mathbf{89.3\pm3.5}$ & $\mathbf{69.3\pm2.4}$ & $\mathbf{67.5\pm4.2}$ \\
\midrule
\modelheader{6}{Llama-3.2-1B-Instruct} \\
--- & Base model
& \multicolumn{2}{c}{$1.8\pm1.3$}
& \multicolumn{2}{c}{$0.3\pm0.4$} \\
Flat & GRPO & $7.8\pm3.5$ & $1.6\pm1.5$ & $1.6\pm0.5$ & $0.0\pm0.0$ \\
Flat & GiGPO & $12.5\pm5.5$ & $7.0\pm2.5$ & $1.6\pm0.3$ & $0.0\pm0.0$ \\
Hier. & HiPER & $28.6\pm5.5$ & $17.2\pm4.4$ & $1.8\pm0.4$ & $0.0\pm0.0$ \\
Hier. & \method{} (ours) & $\mathbf{52.1\pm6.0}$ & $\mathbf{40.4\pm5.0}$ & $\mathbf{17.60\pm3.9}$ & $\mathbf{15.25\pm4.6}$ \\
\bottomrule
\end{tabular}
\end{table*}

Over released HiPER, \method{} gains $+8.1$ points on ALFWorld at the best checkpoint (95\% CI $[+0.2,+16.0]$, $p_{\mathrm{W}}=0.046$) and $+7.8$ at the final checkpoint (CI $[-0.3,+15.9]$, $p_{\mathrm{W}}=0.055$); on WebShop it gains $+7.9$ (CI $[+0.7,+15.1]$, $p_{\mathrm{W}}=0.038$) and $+11.0$ (CI $[+2.4,+19.6]$, $p_{\mathrm{W}}=0.025$).
Three of these four comparisons reach the conventional $0.05$ threshold, the ALFWorld final checkpoint falling just short.
All $p$-values are two-sided unpaired Welch tests with Welch--Satterthwaite degrees of freedom; Table~\ref{tab:welch} collects every comparison, including the flat baselines.
Three-seed inferences are low-powered; intervals, trajectories, and segment-duration analysis are in Appendix~\ref{app:additional-results}.

\paragraph{A second model family.}
Repeating both benchmarks with Llama-3.2-1B-Instruct reproduces the effect at larger magnitude.
On ALFWorld, \method{} reaches $52.1\pm6.0$ best-checkpoint success versus $28.6\pm5.5$ for HiPER ($+23.5$, $p_{\mathrm{W}}=0.0076$) and $40.4\pm5.0$ versus $17.2\pm4.4$ at update 150 ($+23.2$, $p_{\mathrm{W}}=0.0040$).
On WebShop the separation is starker still: \method{} reaches $17.6\pm3.9$ and $15.3\pm4.6$ while HiPER reaches $1.8\pm0.4$ and exactly zero, giving $+15.8$ ($p_{\mathrm{W}}=0.0188$) and $+15.2$ ($p_{\mathrm{W}}=0.0290$).
Every \method{} seed exceeds every HiPER seed on ALFWorld, the maximum rank separation attainable at $n=3$.
Absolute success is far below the Qwen3-0.6B block because this backbone is not instruction-tuned for interactive agents; the relevant quantity is the gap, which widens from $+8.1$ to $+23.5$ points on ALFWorld.
On WebShop the weaker backbone is a discriminative stress setting: HiPER, GRPO, and GiGPO all finish at zero in these runs, so this contrast does not establish a broader ranking of their capability.

\paragraph{Stabilizing temporal abstraction.}
The most direct evidence is behavioral rather than aggregate.
One released HiPER seed ends training with structured-output validity $0.016$, so its boundary decisions become largely unparsable, and its pick-two accuracy stays at $23.3\pm31.8$.
\method{} instead finishes with switch rates $0.150$, $0.266$, and $0.142$, corresponding to multi-turn subgoals, with validity at least $0.996$ on every seed.

\begin{figure*}[t]
\centering
\includegraphics[width=0.9\textwidth]{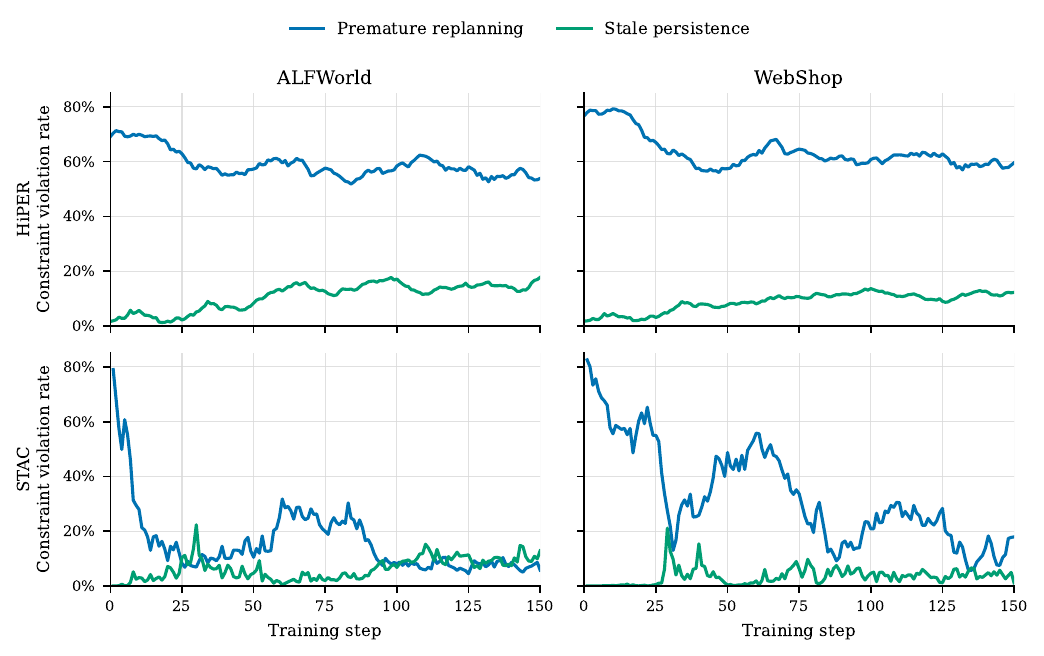}
\caption{\textbf{Constraint violation rates during training.} Per-step violation rates for the two temporal-abstraction constraint surrogates. Each curve averages three seeds. HiPER's premature-replanning rate stays high throughout training, whereas \method{} reduces it and holds both rates low. Because the reported runs use fixed multipliers, these are diagnostics of the learned temporal abstraction rather than enforced target levels.}
\label{fig:constraints}
\end{figure*}

\paragraph{Constraint violation diagnostics.}
Fig.~\ref{fig:constraints} tracks both constraint surrogates throughout training.
Under HiPER, premature replanning stays high, ending at $0.553$ on ALFWorld and $0.597$ on WebShop, so more than half of all turns replace a single-turn subgoal, while stale persistence grows to $0.147$ and $0.110$.
\method{} starts higher still yet drives them to $0.073$ and $0.151$, reductions of $7.6\times$ and $4.0\times$, while also ending lower on stale persistence at $0.098$ and $0.039$; mean segment lengths settle at $4.02$ and $2.91$ turns.
Improving both at once shows the boundary terms do not merely trade one violation for the other.
Because these runs use fixed multipliers, the converged values are the violation levels those prices induce, not budgets the optimizer was asked to meet.
Appendix~\ref{app:ablation} disables one cost at a time and finds each controls its own violation and not the other, an effect resolved at three seeds even though the success differences between arms are not.

\section{Additional Robustness Checks}
\label{sec:robustness}

\paragraph{Fixed penalties, not an adaptive constrained solver.}
The reported duration-based \method{} experiments hold
$\mu_{\mathrm{short}}=\mu_{\mathrm{stale}}=\mu$ fixed throughout training.
They therefore evaluate a fixed, two-sided regularizer on the normalized
boundary advantage rather than the primal--dual update in
Eq.~\eqref{eq:dualupdate}. This objective is related to deliberation-cost
regularization~\cite{harb2018waiting}, but applies separate, mutually
exclusive costs to premature \switch{} and stale \keep{} decisions after the
host has formed its native boundary credit.

\paragraph{Released-backbone replication and multiplier sensitivity.}
To check the local pipeline on the Qwen2.5-1.5B-Instruct backbone used in the
released HiPER study, we ran a matched three-seed ALFWorld comparison with the
same local evaluation and checkpoint protocol. Table~\ref{tab:robustness}
summarizes both robustness checks. Panel~(a) reports a $+3.9$-point
best-checkpoint and $+4.7$-point final-checkpoint gain for the duration-based
method. This is a controlled local comparison, not a claim of seed-matched
reproduction of every upstream aggregate. Panel~(b) varies the shared penalty
on Qwen3-0.6B ALFWorld. Increasing $\mu$ monotonically reduces the measured
violations over the tested range, whereas task success peaks at the
validation-selected $\mu=0.1$ and declines under stronger penalties. We
select $0.1$ on this ALFWorld validation sweep and reuse it unchanged for
WebShop and Llama; this selection coupling is a limitation of the reported
peak values.

\begin{table}[t]
\centering
\caption{\textbf{Robustness checks for duration-based STAC.}
Panel~(a) is a matched local ALFWorld comparison on the Qwen2.5-1.5B-Instruct
HiPER setting. Panel~(b) varies the shared fixed penalty on Qwen3-0.6B
ALFWorld. Values are mean $\pm$ SD over three seeds; lower violation rates
are preferred. Bold marks the best value in each column.}
\label{tab:robustness}
\footnotesize
\setlength{\tabcolsep}{3pt}
\renewcommand{\arraystretch}{1.05}
\begin{tabular}{@{}lcc@{}}
\toprule
\multicolumn{3}{c}{\textit{(a) Qwen2.5-1.5B-Instruct ALFWorld}} \\
\midrule
Method & Best $\uparrow$ & Final $\uparrow$ \\
\midrule
HiPER & $92.5\pm3.7$ & $85.2\pm3.6$ \\
\method{} (duration) & $\mathbf{96.4\pm1.2}$ & $\mathbf{89.9\pm2.1}$ \\
\bottomrule
\end{tabular}

\vspace{0.5em}

\begin{tabular}{@{}lcccc@{}}
\toprule
\multicolumn{5}{c}{\textit{(b) Qwen3-0.6B ALFWorld penalty sensitivity}} \\
\midrule
$\mu$ & Best $\uparrow$ & Final $\uparrow$ & Prem.\ $\downarrow$ & Stale $\downarrow$ \\
\midrule
$0$ & $83.3\pm3.3$ & $81.5\pm3.6$ & $0.553\pm0.041$ & $0.147\pm0.026$ \\
$0.025$ & $87.2\pm3.0$ & $85.1\pm3.2$ & $0.318\pm0.034$ & $0.132\pm0.023$ \\
$0.05$ & $90.1\pm2.6$ & $88.4\pm2.8$ & $0.176\pm0.026$ & $0.111\pm0.020$ \\
$\mathbf{0.10}$ & $\mathbf{91.4\pm3.6}$ & $\mathbf{89.3\pm3.5}$ & $0.073\pm0.018$ & $0.098\pm0.022$ \\
$0.20$ & $90.9\pm2.4$ & $89.0\pm2.6$ & $0.045\pm0.012$ & $0.071\pm0.014$ \\
$0.40$ & $87.5\pm2.9$ & $85.4\pm3.1$ & $\mathbf{0.022\pm0.008}$ & $\mathbf{0.054\pm0.012}$ \\
\bottomrule
\end{tabular}
\end{table}

\paragraph{Scope and remaining limitations.}
For the matched Qwen3 and Llama local HiPER and \method{} arms, we set HiPER's
\texttt{keep\_penalty} and \texttt{keep\_consistency\_penalty} to zero and
retain its environment reward and $-0.1$ malformed-output term. Thus the
comparison does not establish whether a separately tuned host KEEP penalty or
a hard first-turn \switch{} block could recover the same gains. Nor do we
report a $d_{\min}$/$d_{\max}$ sweep: the selected duration values are
pre-specified heuristics, not uniquely justified optima. The short cost
captures most standalone success gain at three seeds, while the stale cost
selectively reduces stale persistence; its incremental task-success benefit
remains unresolved. Broader hosts, more seeds, and these alternative controls
are future work.

\section{Conclusion}

Explicit subgoal syntax alone does not determine how stable the learned temporal abstraction will be.
We identified \emph{temporal abstraction instability} and proposed \method{}, which prices premature replanning and stale persistence as constraint costs on the sampled boundary decision alone, leaving rewards, critic targets, and the subgoal and action advantages untouched.
Over the released hierarchical host it improves best-checkpoint success by $8.1$ and $7.9$ points with Qwen3-0.6B and $23.5$ and $15.8$ points with Llama-3.2-1B-Instruct, seven of these eight comparisons significant at $p_{\mathrm{W}}<0.05$.
The costs are prespecified and the reported runs hold the multipliers fixed; adaptive dual optimization under prescribed budgets is the natural next step.

{
\small
\bibliographystyle{plainnat}
\bibliography{ref}
}

\clearpage
\appendix
\section{Implementation, Experimental Details, and Broader Impact}
\label{app:reproducibility}

\paragraph{Standalone integration.}
\method{} requires a binary boundary decision, a subgoal-age counter, the base boundary advantage, and estimates of the two constraint rates.
It can therefore be inserted after advantage normalization in any on-policy hierarchical optimizer.
The host policy need not use the subgoal text, value heads, or return estimator used in our experiments.
Our experimental structured-output adapter additionally supplies the parsed boundary-token mask described in Appendix~\ref{app:boundary-mask}.
An implementation should expose the constraint-cost functions $g_i$ and the multipliers $\mu_i$, with the budgets $\rho_i$ and a dual optimizer needed only for the adaptive variant; setting both multipliers to zero is the behavior-preserving default for constrained boundary credit.

\begin{figure*}[t]
\centering
\includegraphics[width=0.8\textwidth]{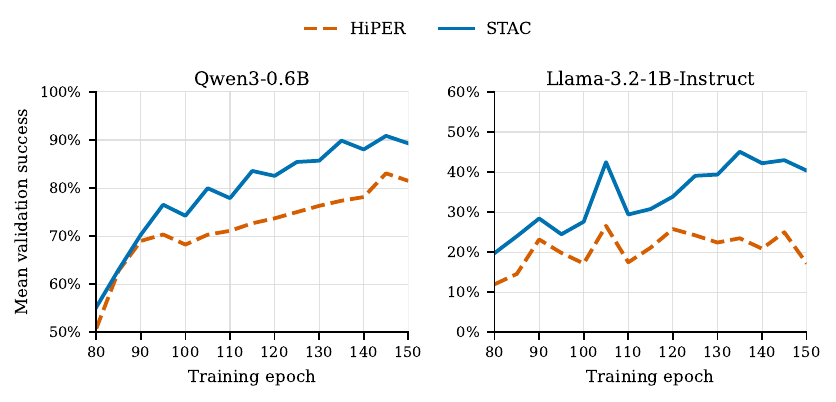}
\caption{\textbf{Late-training ALFWorld validation.} Mean HiPER and \method{} validation success across three seeds from epochs 80 through 150, without confidence intervals. Validation is performed every five epochs on 128 held-out episodes per seed. The vertical ranges differ between panels because the two backbones differ by roughly a factor of two in absolute success; a shared axis would flatten the Llama-3.2-1B panel. \method{} is above HiPER at every evaluation on both backbones over this window.}
\label{fig:curves}
\end{figure*}

\paragraph{Artifacts and statistical protocol.}
Our experiment configuration fixes all non-treatment settings across arms and stores the complete resolved trainer configuration, per-update metrics, validation generations, and sampled training episodes.
We recommend reporting switch rate, segment-length distribution, both constraint violation rates, and valid-format rate alongside task return.
Violation rates and segment statistics are computed only over turns with a parsed valid boundary-token mask; malformed or missing boundary blocks are excluded from those denominators and reported separately through the valid-format rate.
For the main result, we follow HiPER and select each seed's highest-overall-validation checkpoint, then report every category from that checkpoint.
Appendix~\ref{app:final-checkpoint} separately reports step-150 and late-window summaries to expose checkpoint-selection sensitivity.
Figure~\ref{fig:curves} reports the arithmetic mean across the three seed-level validation trajectories without uncertainty intervals.

\paragraph{Training configuration.}
The released ALFWorld host enables the truncation bootstrap in Eq.~\eqref{eq:high-bootstrap}; the WebShop host disables it.
ALFWorld uses a 2048-token prompt limit; WebShop uses 4096 tokens because its released observation template is longer.
The matched Qwen3 and Llama local HiPER and \method{} runs set HiPER's
\texttt{keep\_penalty} and \texttt{keep\_consistency\_penalty} to zero, while
retaining the environment reward and the host's $-0.1$ malformed-output reward
term.
We use all the HiPER baseline numbers in our experiments.
The base-model row evaluates the untrained checkpoint under the same Plan--Execute prompt, temperature, episode limit, and splits.

\paragraph{Experimental Lagrangian instantiation.}
For all reported experiments we use the fixed-multiplier Lagrangian instantiation
\begin{equation}
\mu_{\mathrm{short}}=\mu_{\mathrm{stale}}=0.1,\qquad
g_{\mathrm{short}}=\ind[d_t<2],\quad
g_{\mathrm{stale}}=\ind[d_t\ge8].
\label{eq:experimental-costs}
\end{equation}
For a valid boundary,
$\widehat A_t^{b,\mathrm{STAC}}=\overline A_t^b-0.1(c_t^{\mathrm{short}}+c_t^{\mathrm{stale}})$.
The multipliers are held fixed throughout training rather than updated by Eq.~\eqref{eq:dualupdate}.
The reported result is therefore a fixed boundary penalty, not a primal--dual
method that enforces prescribed violation budgets. We select $0.1$ by
Qwen3-0.6B ALFWorld validation success and reuse it unchanged in the other
reported settings; Table~\ref{tab:robustness} reports the sensitivity
of the selected setting.

\subsection{Heuristic choice of the two duration thresholds}
\label{app:thresholds}

Neither threshold in Eq.~\eqref{eq:experimental-costs} is selected by
validation. They are pre-specified duration heuristics, not guarantees about
the appropriate duration of every individual subgoal.

\paragraph{The lower threshold discourages uniform one-turn collapse.}
The cost $g_{\mathrm{short}}=\ind[d_t<2]$ is active exactly when $d_t=1$.
A subgoal that terminates after a single turn in every state induces the same action distribution as a policy with no subgoal layer at all: the option reduces to a primitive action and the hierarchy is vacuous~\cite{sutton1999options}.
We therefore use $d_{\min}=2$ to discourage this uniform collapse. It does
not imply that every one-turn segment is undesirable, and we do not claim
that $2$ is the only admissible threshold.

\paragraph{The upper threshold is horizon-informed, not a per-segment bound.}
An episode admits at most $T$ environment turns and a task requires $K$ subgoals.
The quantity $T/K$ is an average fair-share reference: if every subgoal
persisted beyond it, the remaining subgoals could not all fit in the horizon.
It is not an upper bound on any individual segment. We use the corresponding
distribution-level reference
\begin{equation}
d_{\max}=8\approx\Bigl\lfloor T/\bar K \Bigr\rfloor ,
\label{eq:fairshare}
\end{equation}
where $\bar K$ is the mean number of subgoals per task in solved benchmark
trajectories.
We estimate $\bar K$ from solved episodes of each benchmark, counting the phases of the recorded plan decomposition, independently of any \method{} run: $\bar K=5.61$ over $113$ ALFWorld episodes and $\bar K=4.62$ over $120$ WebShop episodes.
With $T=50$ this gives $8.91$ on ALFWorld and $10.83$ on WebShop. Choosing
$8$ is therefore a pre-specified, horizon-informed heuristic, not a
mathematical consequence or a tuned optimum.

\paragraph{Observed duration diagnostics.}
Under an identical $d_{\max}=8$, the late-training 90th-percentile segment length is $8.27$ on ALFWorld, sitting at the threshold, and $5.40$ on WebShop, where the agent settles well below it without the constraint being active.
The threshold governs the upper tail of the duration distribution rather than its mean, which remains at $4.02$ and $2.91$ turns.
We report this as a post hoc diagnostic rather than evidence that $8$ is
uniquely correct. A duration-threshold sweep is left to future work.

\paragraph{A per-task adaptive form.}
Because $\bar K$ varies by task category, a task- or state-dependent
$d_{\max}(s_t)=(T-t)/K_{\mathrm{remaining}}$ is a natural extension. We do
not evaluate it here.

\begin{table}[htbp]
\centering
\caption{Horizon fair-share reference $\lfloor T/\bar K\rfloor$ evaluated per ALFWorld task category, with $T=50$ and $\bar K$ measured from solved episodes. The reference is lowest for the most multi-phase category and highest for the fewest-phase category, so Eq.~\eqref{eq:fairshare} is not a constant in disguise.}
\label{tab:fairshare}
\small
\setlength{\tabcolsep}{7pt}
\begin{tabular}{lrrr}
\toprule
Category & Episodes & $\bar K$ & $T/\bar K$ \\
\midrule
\textsc{look}   & $20$ & $3.50$ & $14.3$ \\
\textsc{pick}   & $32$ & $4.09$ & $12.2$ \\
\textsc{clean}  & $17$ & $6.59$ & $7.6$ \\
\textsc{heat}   & $16$ & $6.88$ & $7.3$ \\
\textsc{cool}   & $19$ & $7.11$ & $7.0$ \\
\textsc{pick2}  & $9$  & $8.44$ & $5.9$ \\
\midrule
All             & $113$ & $5.61$ & $8.9$ \\
\bottomrule
\end{tabular}
\end{table}

\paragraph{Compute.}
Every run trains on a single node of eight NVIDIA H100 80GB GPUs, using vLLM for rollout generation and co-locating actor and critic on the same node.
A 150-update run takes roughly 12--24 hours depending on benchmark and backbone, WebShop being slower than ALFWorld because of its longer observation template.

\paragraph{Broader impact.}
Temporal abstraction can make language agents more efficient and interpretable by exposing the subgoal that persists across actions.
However, the chosen constraint-cost functions encode a behavioral prior.
Costs that are too restrictive can suppress legitimate rapid replanning or force stale subgoals, especially under distribution shift.
For deployment, boundary credit should be monitored, capped, and disabled when the structured interface is not semantically validated.
It does not replace task-level safety constraints.

\section{Host Advantage Estimator}
\label{app:host-advantage}

HiPER uses one shared critic backbone with two scalar output heads~\cite{peng2026hiper}.
The high-level head $V_\phi^H(s_t)$ estimates the return from selecting a new subgoal at a segment boundary, while the low-level head $V_\phi^L(s_t,z_t)$ estimates the return from continuing execution under the active subgoal.
There is no separately learned boundary-value head.
The two heads produce three actor advantages because the boundary advantage is derived from their value difference.

Let $0=b_0<b_1<\cdots<b_K=T$ be the segment boundaries.
Segment $k$ contains turns $[b_k,b_{k+1}-1]$, has duration $D_k=b_{k+1}-b_k$, and uses subgoal $z_k$.
Write $\operatorname{sg}[\cdot]$ for stop-gradient and define
\begin{equation}
B_{k+1}^H=
\begin{cases}
V_\phi^H(s_{b_{k+1}}), & b_{k+1}<T,\\
\operatorname{sg}[V_\phi^H(s_{b_k})], & \text{bootstrapped truncation},\\
0, & \text{otherwise}.
\end{cases}
\label{eq:high-bootstrap}
\end{equation}
The middle case is the released host's optional treatment of a trajectory cut off by the turn limit.
The discounted segment reward, high-level residual, and subgoal advantage are
\begin{align}
\widetilde r_k&=\sum_{j=0}^{D_k-1}\gamma^j r_{b_k+j}, \notag\\
\delta_k^z&=\widetilde r_k+\gamma^{D_k}B_{k+1}^H-V_\phi^H(s_{b_k}), \notag\\
\widehat A_{b_k}^z&=\delta_k^z+\gamma^{D_k}\lambda_H\widehat A_{b_{k+1}}^z.
\label{eq:high-advantage}
\end{align}
$\widehat A_{b_k}^z$ is defined at segment starts and is assigned to the newly generated subgoal tokens.

Within a segment, define
\begin{equation}
\widetilde V_{t+1}=
\begin{cases}
V_\phi^L(s_{t+1},z_k), & t<b_{k+1}-1,\\
B_{k+1}^H, & t=b_{k+1}-1.
\end{cases}
\end{equation}
The low-level residual and action advantage are
\begin{align}
\delta_t^a&=r_t+\gamma\widetilde V_{t+1}-V_\phi^L(s_t,z_k), \notag\\
\widehat A_t^a
&=\sum_{\ell=t}^{b_{k+1}-1}
(\gamma\lambda_L)^{\ell-t}\delta_\ell^a,
\qquad t\in[b_k,b_{k+1}-1].
\label{eq:low-advantage}
\end{align}
The sum stops at the segment boundary, where the final residual bootstraps to the high-level head.
$\widehat A_t^a$ is assigned to action tokens.

For boundary credit, define the estimated gain from replacing the previous subgoal,
\begin{align}
\Delta_t^b
&=V_\phi^H(s_t)-V_\phi^L(s_t,z_{t-1}), \notag\\
\widehat A_t^b
&=(y_t-\beta_t)\Delta_t^b,
\quad
y_t=\ind[q_t=\qswitch],\quad
\beta_t=\pi_\theta^b(\qswitch\mid s_t,z_{t-1}).
\label{eq:boundary-advantage}
\end{align}
This is a centered policy-gradient estimator for the realized binary decision and is assigned to the boundary tokens at every turn.
Before the actor update, the advantage streams are independently standardized.
The boundary stream gives the normalized boundary advantage $\overline A_t^b$ of Eq.~\eqref{eq:normalized-boundary-advantage}, whose batch $\mathcal B$ contains exactly those turns with a valid boundary-token mask.
This $\overline A_t^b$ is the sole host advantage consumed by \method{}.
The two critic targets are
$y_k^H=\widetilde r_k+\gamma^{D_k}\operatorname{sg}[B_{k+1}^H]$ and
$y_t^L=r_t+\gamma\operatorname{sg}[\widetilde V_{t+1}]$.
The critic regresses $V_\phi^H(s_{b_k})$ to $y_k^H$ and $V_\phi^L(s_t,z_k)$ to $y_t^L$.
\method{} changes neither target.

\section{Optimization Details}
\label{app:optimization-details}

\paragraph{Locality.}
Potential-based reward shaping~\cite{ng1999shaping} replaces the trajectory reward $r_t$ with
$r_t'=r_t+\gamma\Phi(s_{t+1})-\Phi(s_t)$.
Because bootstrapped estimators aggregate discounted rewards through the critic, such a change can alter critic targets and advantages at earlier turns.
\method{} instead adds the sampled Lagrangian cost terms after host advantage estimation, only to the boundary span at turn $t$.
Thus it changes neither the reward stream, critic targets, nor another turn's advantage.
Moreover, the mutually exclusive boundary decisions and bounded costs imply that the valid-decision perturbation is bounded by
$\max(\mu_{\mathrm{short}},\mu_{\mathrm{stale}})$.

\paragraph{One update.}
Each update (1) rolls out the hierarchical policy, (2) tracks $d_t$ and computes Eq.~\eqref{eq:violation},
(3) computes and normalizes the host advantage streams, (4) forms Eq.~\eqref{eq:auxadv},
and (5) applies the unchanged on-policy actor and critic updates.
Step (6) applies only to the adaptive primal--dual variant, which additionally updates the multipliers using Eq.~\eqref{eq:dualupdate}; the reported runs omit it and hold $\boldsymbol\mu$ fixed.
We log switch rate, segment-length quantiles, one-turn-segment rate, and both constraint rates, together with the multipliers whenever they are adapted.

\section{Classical Termination Regularization}

Classical hierarchical reinforcement learning represents the keep/switch decision as a termination function.
Option-Critic learns termination policies end to end~\cite{bacon2017option}, while deliberation costs discourage frequent termination~\cite{harb2018waiting}.
A one-sided termination cost can favor indefinitely long options, and both formulations assume a closed, indexed decision space.
\method{} instead constrains both premature replanning and stale persistence at the boundary-policy level.
It is compatible with GAE~\cite{schulman2016gae}, PPO~\cite{schulman2017ppo}, and other on-policy hierarchical objectives.

\section{Related Hierarchical Language-Agent Methods}
\label{app:related-hrl-agents}

Recent hierarchical language-agent methods differ in what constitutes a level of the hierarchy, how those levels are trained, and where credit is assigned.
This section summarizes the methods cited in the introduction from their primary papers and official implementations.

\paragraph{ArCHer.}
ArCHer is a two-level actor--critic framework for multi-turn language-model interaction~\cite{zhou2024archer}.
Its high level applies off-policy temporal-difference learning to utterance- or turn-level transitions, thereby propagating reward across an entire dialogue, while its low level applies token-level policy optimization within each utterance using the high-level value function for cross-turn credit.
This separation allows the high and low levels to run RL algorithms with different update rules and substantially improves sample efficiency relative to fully on-policy multi-turn optimization.

\paragraph{GLIDER.}
GLIDER uses a two-level planner--controller hierarchy for grounding language models in sequential environments~\cite{hu2025glider}.
The high-level policy generates abstract chain-of-thought plans or subtasks, while a parameter-efficient low-level controller executes environment actions conditioned on the current plan.
Training proceeds from behavior-cloning initialization to offline hierarchical RL with an advantage-weighted actor--critic objective, followed by offline-to-online adaptation; the paper evaluates transfer and generalization on ScienceWorld and ALFWorld.

\paragraph{STEP-HRL.}
STEP-HRL is the method name used in the paper formally titled \emph{Hierarchical Reinforcement Learning with Augmented Step-Level Transitions for LLM Agents}~\cite{zhen2026stephrl}.
It introduces three functional components: a high-level policy that predicts the next subtask from the instruction and completed-subtask history, a low-level policy that acts from the current observation, subgoal, and compact local progress, and a local-progress module that iteratively folds intra-subtask interactions into that compact representation.
The components share policy parameters but use distinct critics; training combines behavior-cloning initialization with offline RL.
The compact progress state reduces dependence on full interaction histories and is evaluated on ScienceWorld and ALFWorld.

\paragraph{HiMAC.}
HiMAC decomposes control into a shared-parameter Macro-Policy and Micro-Policy~\cite{jin2026himac}.
The Macro-Policy generates a structured blueprint of natural-language subgoals, and the Micro-Policy generates atomic actions conditioned on each subgoal.
Its critic-free hierarchical optimization extends group-relative policy optimization to produce separate blueprint-level and execution-level relative advantages.
An alternating co-evolution schedule updates macro exploration with the executor fixed, then adapts the executor under a fixed high-confidence blueprint; experiments cover ALFWorld, WebShop, and Sokoban.

\paragraph{HIPIF.}
HIPIF combines hierarchical planning with information folding for long-horizon interaction~\cite{diao2026hipif}.
At each step, a reflection operation decides whether the active subgoal is complete.
When complete, the method folds that subgoal's interaction history into a compact summary and generates the next subgoal; otherwise, it generates another primitive action under the current subgoal.
Training uses a GRPO-style clipped objective whose step-level advantages combine terminal environment reward with subgoal-oriented process rewards, avoiding an ever-growing raw interaction context.

\paragraph{Multi\texorpdfstring{$^2$}{2}.}
Multi$^2$ assigns the two hierarchy levels to separate role-specialized agents~\cite{park2026multi2}.
Its high-level System~1 agent generates context-aware subgoals and is trained by supervised fine-tuning on a subgoal dataset.
Its low-level System~2 agent executes atomic actions, first learning from offline data with actor--critic RL and then receiving online RL fine-tuning while System~1 remains fixed.
The method targets objective drift in long-horizon interaction and introduces hierarchical datasets spanning multiple interactive environments.

\paragraph{DreamPhase.}
A complementary line of work improves planning quality rather than hierarchical structure.
DreamPhase equips a language agent with offline imagination, planning over internally generated rollouts rather than additional environment interaction, and uses uncertainty estimates to guide which plan to follow~\cite{hamidi2026dreamphase}.
This targets \emph{what} the agent should plan, whereas \method{} targets \emph{when} an active subgoal should be replaced, so the two are compatible: \method{} constrains only the boundary decision of whichever planner is in use.

\paragraph{Relation to STAC.}
The hierarchical methods above demonstrate several benefits of hierarchical language-agent training, including cross-turn credit assignment, transferable low-level skills, compact progress representations, macro--micro optimization, folded histories, and role-specialized planning and execution.
STAC addresses an orthogonal issue: regardless of how a host defines or trains its hierarchy, the learned boundary can still switch too frequently, persist after becoming stale, or leave the parsed optimization support.
It therefore consumes a host-provided boundary advantage and modifies only boundary credit rather than prescribing a new planner, executor, trajectory representation, or base return estimator.

\section{Additional Results}
\label{app:additional-results}

\begin{figure*}[t]
\centering
\includegraphics[width=\textwidth]{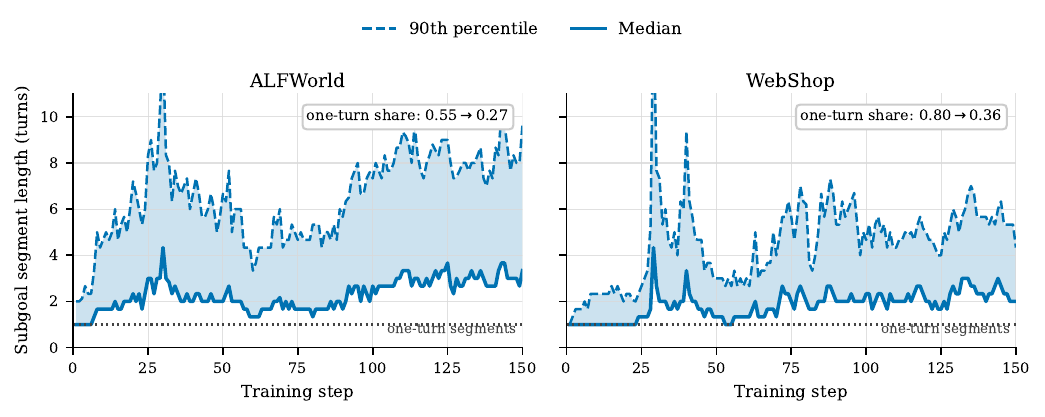}
\caption{\textbf{Subgoal durations move away from one-turn segments.} Median and 90th-percentile \method{} segment length, each computed over all segments in a training update and then averaged across three seeds; the shaded region spans the 50th--90th percentile range. The annotation gives the fraction of one-turn segments over the first twenty and final thirty updates.}
\label{fig:durations}
\end{figure*}

\paragraph{Duration distribution.}
Switch rate alone summarizes temporal abstraction with a single scalar, so Fig.~\ref{fig:durations} reports how the segment-length distribution itself evolves under \method{}.
Early in training the distribution concentrates on one-turn segments: they account for $0.55$ of ALFWorld segments and $0.80$ of WebShop segments, and the WebShop median segment is exactly one turn.
By the end of training that mass has roughly halved, to $0.27$ and $0.36$, and the distribution has both shifted and widened.
The median rises from $1.55$ to $3.03$ turns on ALFWorld and from $1.00$ to $2.30$ on WebShop, while the 90th percentile rises from $4.27$ to $8.27$ and from $2.08$ to $5.40$.
The upper tail therefore grows faster than the median, which is the intended behavior: subgoals that coordinate long action sequences persist, while short segments remain available when replanning is appropriate.

\paragraph{Base model.}
Before RL training, Qwen3-0.6B succeeds on $1/384$ ALFWorld episodes and $3/256$ WebShop episodes, yielding $0.3\pm0.5\%$ and $1.2\pm1.7\%$ success.
Its graded WebShop task score is $6.0\pm1.9\%$; Table~\ref{tab:results} uses binary success consistently.

\begin{figure*}[t]
\centering
\includegraphics[width=\textwidth]{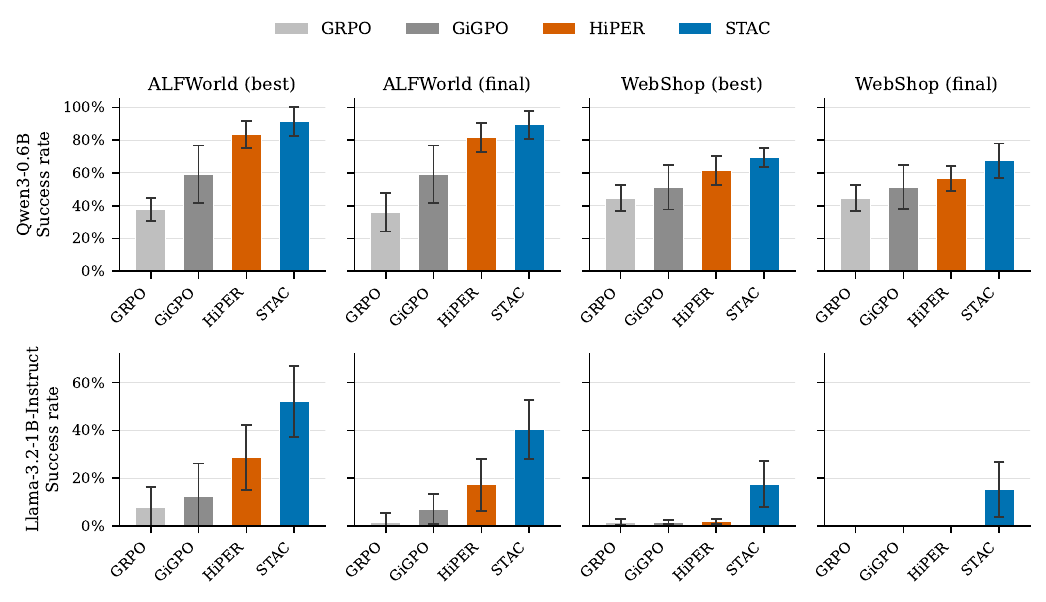}
\caption{\textbf{Main results with per-mean confidence intervals.} The values of Table~\ref{tab:results} plotted as success rates. Error bars are $95\%$ Student-$t$ confidence intervals for each mean, computed as $t_{0.975,2}\,s/\sqrt{3}$ from the three-seed standard deviation, so they describe the precision of a single arm rather than the significance of a difference. At $n=3$ the multiplier is $t_{0.975,2}=4.30$, which is why the intervals are wide and frequently overlap even where the Welch test in Table~\ref{tab:welch} rejects. On Llama-3.2-1B-Instruct WebShop at the final checkpoint, every baseline records exactly zero across seeds and therefore has no interval.}
\label{fig:results-bars}
\end{figure*}

\paragraph{Statistical comparisons.}
Fig.~\ref{fig:results-bars} visualizes Table~\ref{tab:results} with a confidence interval for each individual mean.
Those intervals overlap in most panels, which reflects the width of a three-seed $t$ interval rather than an absence of separation; the comparisons in Table~\ref{tab:welch} test the difference directly and are the quantity of interest.
Table~\ref{tab:welch} reports every method against released HiPER as a two-sided unpaired Welch test with Welch--Satterthwaite degrees of freedom, computed from the three-seed means and standard deviations printed in Table~\ref{tab:results} rather than from per-seed values.
\method{} improves on HiPER in all eight benchmark--checkpoint cells, and seven of the eight reach $p_{\mathrm{W}}<0.05$; the exception is the Qwen3-0.6B ALFWorld final checkpoint at $p_{\mathrm{W}}=0.0547$.
Both flat baselines are below HiPER wherever the comparison is defined, so the hierarchy itself helps and the boundary terms add a further gain on top of it.
Two limitations bound these numbers.
First, every test uses three seeds, so the degrees of freedom are between two and four and the intervals are wide.
Second, on Llama-3.2-1B-Instruct at the final checkpoint, HiPER, GRPO, and GiGPO all record $0.0\pm0.0$, so the flat-baseline comparisons have zero difference and zero pooled variance and no Welch statistic exists; those cells are marked ``---'' rather than assigned a $p$-value.
Deriving tests from published means and standard deviations also discards the seed-level pairing, which is the conservative choice here.

\begin{table}[t]
\centering
\caption{Two-sided unpaired Welch tests against released HiPER. Differences are in success points; $^{*}$ denotes $p_{\mathrm{W}}<0.05$.}
\label{tab:welch}
\small
\setlength{\tabcolsep}{4pt}
\begin{tabular}{llrrrr}
\toprule
Benchmark & Method & Diff. & $t$ & 95\% CI & $p_{\mathrm{W}}$ \\
\midrule
\modelheader{6}{Qwen3-0.6B} \\
ALFWorld best & GRPO & $-45.8$ & $-18.33$ & $[-52.8,-38.8]$ & $0.0001^{*}$ \\
              & GiGPO & $-24.2$ & $-5.35$ & $[-39.1,-9.3]$ & $0.0148^{*}$ \\
              & \method{} & $\mathbf{+8.1}$ & $2.87$ & $[+0.3,+16.0]$ & $\mathbf{0.0457^{*}}$ \\
ALFWorld final & GRPO & $-45.6$ & $-13.34$ & $[-55.4,-35.9]$ & $0.0003^{*}$ \\
              & GiGPO & $-22.4$ & $-4.87$ & $[-37.1,-7.7]$ & $0.0170^{*}$ \\
              & \method{} & $\mathbf{+7.8}$ & $2.69$ & $[-0.3,+15.9]$ & $0.0547$ \\
WebShop best & GRPO & $-16.7$ & $-6.10$ & $[-24.3,-9.1]$ & $0.0038^{*}$ \\
              & GiGPO & $-10.1$ & $-2.68$ & $[-21.3,+1.1]$ & $0.0656$ \\
              & \method{} & $\mathbf{+7.9}$ & $3.22$ & $[+0.7,+15.1]$ & $\mathbf{0.0383^{*}}$ \\
WebShop final & GRPO & $-11.8$ & $-4.66$ & $[-18.8,-4.8]$ & $0.0097^{*}$ \\
              & GiGPO & $-5.2$ & $-1.46$ & $[-16.3,+5.9]$ & $0.2374$ \\
              & \method{} & $\mathbf{+11.0}$ & $3.69$ & $[+2.4,+19.6]$ & $\mathbf{0.0250^{*}}$ \\
\midrule
\modelheader{6}{Llama-3.2-1B-Instruct} \\
ALFWorld best & GRPO & $-20.8$ & $-5.53$ & $[-32.0,-9.6]$ & $0.0084^{*}$ \\
              & GiGPO & $-16.1$ & $-3.59$ & $[-28.6,-3.6]$ & $0.0231^{*}$ \\
              & \method{} & $\mathbf{+23.5}$ & $5.00$ & $[+10.4,+36.6]$ & $\mathbf{0.0076^{*}}$ \\
ALFWorld final & GRPO & $-15.6$ & $-5.81$ & $[-25.3,-5.9]$ & $0.0172^{*}$ \\
              & GiGPO & $-10.2$ & $-3.49$ & $[-19.2,-1.2]$ & $0.0364^{*}$ \\
              & \method{} & $\mathbf{+23.2}$ & $6.03$ & $[+12.5,+33.9]$ & $\mathbf{0.0040^{*}}$ \\
WebShop best & GRPO & $-0.2$ & $-0.54$ & $[-1.3,+0.9]$ & $0.6185$ \\
              & GiGPO & $-0.2$ & $-0.69$ & $[-1.0,+0.6]$ & $0.5293$ \\
              & \method{} & $\mathbf{+15.8}$ & $6.98$ & $[+6.3,+25.4]$ & $\mathbf{0.0188^{*}}$ \\
WebShop final & GRPO & $0.0$ & --- & --- & --- \\
              & GiGPO & $0.0$ & --- & --- & --- \\
              & \method{} & $\mathbf{+15.2}$ & $5.74$ & $[+3.8,+26.7]$ & $\mathbf{0.0290^{*}}$ \\
\bottomrule
\end{tabular}
\end{table}

\paragraph{Multi-phase behavior.}
At selected checkpoints, pick-two success rises from $0.231\pm0.310$ to $0.811\pm0.078$; the wide HiPER spread reflects seed-level bimodality rather than a stable mean.
The largest observed gain is concentrated in this multi-phase category.

\paragraph{Hierarchy validity.}
All \method{} seeds retain multi-turn subgoals, nearly perfect structured-output validity, and no final missing boundary masks.
One released HiPER seed instead ends at validity $0.016$ despite validation success $0.812$, so aggregate accuracy alone does not reveal the state of the hierarchy.

\section{Constraint Ablation}
\label{app:ablation}

\method{} adds two costs, and it is natural to ask how much each contributes.
We isolate them by disabling one constraint at a time on Qwen3-0.6B, holding every other
setting fixed: \emph{short only} keeps $g_{\mathrm{short}}=\ind[d_t<2]$ and removes the
stale cost, \emph{stale only} keeps $g_{\mathrm{stale}}=\ind[d_t\ge8]$ and removes the
short cost.
Both reduce to HiPER when the remaining multiplier is also set to zero.
Tables~\ref{tab:ablation-alfworld} and~\ref{tab:ablation-webshop} report the result.

\paragraph{Each cost controls its own violation.}
The two constraints are selective in the expected direction.
Adding the short cost alone drives premature replanning from $0.553$ to $0.085$ on ALFWorld
and from $0.597$ to $0.165$ on WebShop, while leaving stale persistence uncontrolled.
Adding the stale cost alone drives stale persistence from $0.147$ to $0.045$ and from
$0.110$ to $0.025$, the lowest stale rates in either table, while leaving premature
replanning essentially at its HiPER value ($0.565$ and $0.605$).
Neither cost substitutes for the other.

\paragraph{The two failure modes are not symmetric.}
By point estimate, short only recovers most of the gain, reaching $89.2$ of \method{}'s
$91.4$ on ALFWorld and $67.4$ of $69.3$ on WebShop, whereas stale only sits within noise of
HiPER ($84.1$ and $62.0$ against $83.3$ and $61.4$; $p_{\mathrm{W}}=0.79$ and $0.84$).
This ordering suggests that premature replanning is the dominant failure mode at this scale:
it is by far the more frequent violation under HiPER, and the arm that suppresses it is the
arm that moves success.
Suppressing stale persistence alone corrects a rarer failure and leaves subgoals that still
collapse to single turns, which is visible in the mean segment length of $1.6$ turns,
essentially HiPER's $1.7$.

\paragraph{Why both are still needed.}
Each single-constraint arm leaves the violation it does not control no better than HiPER,
and the short-only arm trends worse.
Its stale-persistence rate is $0.185\pm0.031$ against HiPER's $0.147\pm0.026$, a gap that
is directional rather than resolved at $n=3$ ($p_{\mathrm{W}}=0.18$), and its segments
lengthen to $4.70\pm0.61$ turns, consistent with pressure away from switching acting as
pressure toward keeping.
The full method attains the lower stale rate of the two ($0.098\pm0.022$) at a shorter
$4.02\pm0.47$ turns, which is what we mean by the stale cost acting as a brake; we note
that this particular contrast is suggestive rather than established at this seed count.
Multi-phase Pick2 accuracy follows the same ordering, rising from $23.1$ under HiPER to
$68.5$ with the short cost, $27.5$ with the stale cost, and $81.1$ with both, consistent
with the claim that this category requires holding one subgoal across an entire
fetch-and-place phase.

\paragraph{What three seeds do and do not resolve.}
The violation effects are large relative to their spread and are resolved at this seed
count: the short cost reduces premature replanning from $0.553\pm0.041$ to
$0.085\pm0.021$ on ALFWorld ($p_{\mathrm{W}}=0.0004$) and from $0.597\pm0.034$ to
$0.165\pm0.029$ on WebShop ($p_{\mathrm{W}}=0.0001$), while the stale cost reduces stale
persistence from $0.147\pm0.026$ to $0.045\pm0.013$ ($p_{\mathrm{W}}=0.0095$).
The success differences between arms are not: short only versus HiPER gives
$p_{\mathrm{W}}=0.11$ on ALFWorld and $0.087$ on WebShop, and full versus short only gives
$p_{\mathrm{W}}=0.51$.
The Pick2 column is the least resolved of all, because HiPER and the stale-only arm are
seed-bimodal with standard deviations of $31.0$ and $28.4$.
We therefore read this ablation as establishing \emph{which cost controls which violation},
and as ordering the arms by point estimate, rather than as separating their task success.

\begin{table}[htbp]
\centering
\caption{\textbf{Constraint ablation on ALFWorld} (Qwen3-0.6B). ``Short'' and ``Stale'' mark which constraint is active; the HiPER row is the host with both multipliers at zero and the \method{} row is the configuration used in Table~\ref{tab:results}. ``(short)'' and ``(stale)'' are the single-constraint arms and ``(full)'' is \method{} as reported elsewhere in the paper. Values are mean $\pm$ SD across three seeds; violation rates and mean segment length are averaged over the final thirty updates. Arrows give the preferred direction. Bold marks the best value in each column. At $n=3$ the violation-rate differences are resolved but the success differences between arms are not; see the text.}
\label{tab:ablation-alfworld}
\footnotesize
\setlength{\tabcolsep}{2.5pt}
\begin{tabular}{lccrrrrrr}
\toprule
Method & Short & Stale & Best $\uparrow$ & Final $\uparrow$ & Prem.\ $\downarrow$ & Stale $\downarrow$ & Seg.\ len. & Pick2 $\uparrow$ \\
\midrule
HiPER                & $\times$     & $\times$     & $83.3\pm3.3$ & $81.5\pm3.6$ & $0.553\pm0.041$ & $0.147\pm0.026$ & $1.70\pm0.24$ & $23.1\pm31.0$ \\
\method{} (short)   & $\checkmark$ & $\times$     & $89.2\pm3.8$ & $87.0\pm3.9$ & $0.085\pm0.021$ & $0.185\pm0.031$ & $4.70\pm0.61$ & $68.5\pm14.6$ \\
\method{} (stale)   & $\times$     & $\checkmark$ & $84.1\pm3.5$ & $82.4\pm3.7$ & $0.565\pm0.039$ & $\mathbf{0.045\pm0.013}$ & $1.60\pm0.22$ & $27.5\pm28.4$ \\
\method{} (full)    & $\checkmark$ & $\checkmark$ & $\mathbf{91.4\pm3.6}$ & $\mathbf{89.3\pm3.5}$ & $\mathbf{0.073\pm0.018}$ & $0.098\pm0.022$ & $4.02\pm0.47$ & $\mathbf{81.1\pm7.8}$ \\
\bottomrule
\end{tabular}
\end{table}

\begin{table}[htbp]
\centering
\caption{\textbf{Constraint ablation on WebShop} (Qwen3-0.6B). Columns follow Table~\ref{tab:ablation-alfworld}. The same ordering holds: the short cost carries most of the success gain, the stale cost gives the lowest stale-persistence rate, and only the full method is low on both.}
\label{tab:ablation-webshop}
\small
\setlength{\tabcolsep}{5pt}
\begin{tabular}{lccrrrr}
\toprule
Method & Short & Stale & Best $\uparrow$ & Final $\uparrow$ & Premature $\downarrow$ & Stale $\downarrow$ \\
\midrule
HiPER                & $\times$     & $\times$     & $61.4\pm3.5$ & $56.5\pm3.0$ & $0.597\pm0.034$ & $0.110\pm0.020$ \\
\method{} (short)    & $\checkmark$ & $\times$     & $67.4\pm2.9$ & $64.3\pm3.7$ & $0.165\pm0.029$ & $0.095\pm0.019$ \\
\method{} (stale)    & $\times$     & $\checkmark$ & $62.0\pm3.4$ & $57.4\pm3.2$ & $0.605\pm0.036$ & $\mathbf{0.025\pm0.009}$ \\
\method{} (full)     & $\checkmark$ & $\checkmark$ & $\mathbf{69.3\pm2.4}$ & $\mathbf{67.5\pm4.2}$ & $\mathbf{0.151\pm0.027}$ & $0.039\pm0.011$ \\
\bottomrule
\end{tabular}
\end{table}

\section{Final-Checkpoint and Late-Window Robustness}
\label{app:final-checkpoint}

The main text follows HiPER's best-validation-checkpoint convention.
For completeness, Table~\ref{tab:bestcategories} reports the ALFWorld category breakdown at the saved best checkpoints, Table~\ref{tab:finalcategories} reports every category at the final checkpoint, and Table~\ref{tab:latecategories} gives a lower-variance late-window diagnostic.
``Late'' first averages the seven validations at updates 120--150 within each seed and then reports mean $\pm$ SD across seeds.
Tables~\ref{tab:bestcategories} and~\ref{tab:finalcategories} additionally carry a Llama-3.2-1B-Instruct block; the late-window diagnostic in Table~\ref{tab:latecategories} remains Qwen3-0.6B only, since the flat baselines were not rerun on the second backbone.

\paragraph{Reading the Llama-3.2-1B block.}
This backbone never solves the two appliance categories: Heat is $0.0$ for every seed of both methods, and Cool is $0.0$ everywhere except one \method{} seed at the final checkpoint.
Pick2 is the diagnostic category, because completing it requires holding one subgoal across a full fetch-and-place phase before switching to the second object.
There \method{} reaches $80.1\pm12.8$ against HiPER's $20.1\pm9.4$ at the best checkpoint and $61.7\pm16.9$ against $8.3\pm10.4$ at update 150, mirroring the same Pick2 weakness that HiPER shows on Qwen3-0.6B ($23.1\pm31.0$).

\clearpage
\begin{table*}[t]
\centering
\caption{ALFWorld category accuracy (\%) at the saved best checkpoints used in Table~\ref{tab:results}, mean $\pm$ SD across three seeds. ``All'' is logged task-distribution-weighted overall accuracy. Bold marks the better method within each model block.}
\label{tab:bestcategories}
\small
\resizebox{\textwidth}{!}{%
\begin{tabular}{lrrrrrrr}
\toprule
Method & Pick & Look & Clean & Heat & Cool & Pick2 & All \\
\midrule
\modelheader{8}{Qwen3-0.6B} \\
\quad GRPO
& $51.0\pm3.4$ & $21.0\pm18.3$ & $43.9\pm9.2$ & $31.9\pm15.5$
& $25.3\pm17.6$ & $40.7\pm1.2$ & $37.5\pm2.8$ \\
\quad GiGPO
& $75.1\pm12.0$ & $55.6\pm12.7$ & $57.9\pm13.9$ & $50.1\pm10.3$
& $53.3\pm8.3$ & $55.0\pm10.0$ & $59.1\pm7.1$ \\
\quad HiPER
& $\mathbf{98.1\pm1.6}$ & $\mathbf{85.8\pm19.5}$ & $95.6\pm5.1$ & $97.0\pm5.3$
& $\mathbf{86.5\pm9.2}$ & $23.1\pm31.0$ & $83.3\pm3.3$ \\
\quad \method{} (ours)
& $94.9\pm8.9$ & $\mathbf{85.8\pm4.7}$ & $\mathbf{97.5\pm2.2}$ & $\mathbf{97.8\pm3.9}$
& $85.6\pm11.2$ & $\mathbf{81.1\pm7.8}$ & $\mathbf{91.4\pm3.6}$ \\
\midrule
\modelheader{8}{Llama-3.2-1B-Instruct} \\
\quad HiPER
& $61.5\pm11.4$ & $65.9\pm7.3$ & $0.0\pm0.0$ & $0.0\pm0.0$
& $0.0\pm0.0$ & $20.1\pm9.4$ & $28.6\pm5.5$ \\
\quad \method{} (ours)
& $\mathbf{86.4\pm10.0}$ & $\mathbf{70.8\pm6.6}$ & $\mathbf{29.0\pm10.3}$ & $0.0\pm0.0$
& $0.0\pm0.0$ & $\mathbf{80.1\pm12.8}$ & $\mathbf{52.1\pm6.0}$ \\
\bottomrule
\end{tabular}
}
\end{table*}

\begin{table*}[t]
\centering
\caption{Final-checkpoint category accuracy (\%) at update 150, mean $\pm$ SD across three seeds. ``Avg.'' is logged task-distribution-weighted overall accuracy. Bold marks the better method within each model block.}
\label{tab:finalcategories}
\small
\resizebox{\textwidth}{!}{%
\begin{tabular}{lrrrrrrr}
\toprule
Method & Pick & Look & Clean & Heat & Cool & Pick2 & Avg. \\
\midrule
\modelheader{8}{Qwen3-0.6B} \\
\quad GRPO
& $48.9\pm8.9$ & $27.8\pm12.7$ & $47.4\pm5.3$ & $14.1\pm0.4$
& $40.0\pm6.9$ & $28.3\pm12.6$ & $35.9\pm4.7$ \\
\quad GiGPO
& $75.1\pm12.0$ & $55.6\pm12.7$ & $57.9\pm13.9$ & $50.1\pm10.3$
& $53.3\pm8.3$ & $55.0\pm10.0$ & $59.1\pm7.1$ \\
\quad HiPER
& $96.8\pm5.6$ & $83.3\pm0.0$ & $\mathbf{98.2\pm3.1}$ & $\mathbf{92.2\pm7.2}$
& $86.7\pm2.3$ & $23.3\pm31.8$ & $81.5\pm3.6$ \\
\quad \method{}
& $\mathbf{97.8\pm3.8}$ & $\mathbf{91.7\pm8.4}$ & $93.0\pm8.0$ & $84.6\pm10.3$
& $\mathbf{92.0\pm6.9}$ & $\mathbf{73.3\pm17.6}$ & $\mathbf{89.3\pm3.5}$ \\
\midrule
\modelheader{8}{Llama-3.2-1B-Instruct} \\
\quad HiPER
& $47.7\pm9.8$ & $47.2\pm7.6$ & $0.0\pm0.0$ & $0.0\pm0.0$
& $0.0\pm0.0$ & $8.3\pm10.4$ & $17.2\pm4.4$ \\
\quad \method{}
& $\mathbf{86.0\pm15.9}$ & $\mathbf{72.2\pm11.0}$ & $\mathbf{19.3\pm13.4}$ & $0.0\pm0.0$
& $\mathbf{2.7\pm4.6}$ & $\mathbf{61.7\pm16.9}$ & $\mathbf{40.4\pm5.0}$ \\
\bottomrule
\end{tabular}
}
\end{table*}

Relative to HiPER, STAC's final-overall unpaired Welch difference is $+0.078$ (95\% CI $[-0.003,+0.159]$, $p_{\mathrm{W}}=0.0547$).
Its late-overall difference, averaging updates 120--150, is $+0.095$ ($[+0.052,+0.138]$).
These parametric intervals remain preliminary at three seeds.
The final accuracies are $0.359\pm0.047$ for GRPO, $0.591\pm0.071$ for GiGPO,
$0.815\pm0.036$ for HiPER, and $0.893\pm0.035$ for STAC.

\begin{table*}[t]
\centering
\caption{Late-window per-category accuracy (\%), mean $\pm$ SD across seeds. ``Avg.'' is logged task-distribution-weighted overall accuracy.}
\label{tab:latecategories}
\small
\resizebox{\textwidth}{!}{%
\begin{tabular}{lrrrrrrr}
\toprule
Method & Pick & Look & Clean & Heat & Cool & Pick2 & Avg. \\
\midrule
GRPO
& $39.4\pm4.6$ & $23.5\pm9.1$ & $32.8\pm6.3$ & $28.3\pm3.1$
& $18.3\pm1.1$ & $20.7\pm3.5$ & $28.5\pm1.5$ \\
GiGPO
& $61.6\pm9.3$ & $46.7\pm8.9$ & $56.0\pm11.0$ & $50.1\pm9.7$
& $29.0\pm4.1$ & $38.8\pm9.6$ & $48.2\pm7.9$ \\
HiPER
& $\mathbf{96.7\pm1.3}$ & $77.4\pm1.5$ & $92.4\pm0.6$ & $86.2\pm7.7$
& $85.0\pm0.4$ & $11.6\pm13.3$ & $77.9\pm2.0$ \\
\method{}
& $94.9\pm2.0$ & $\mathbf{81.6\pm2.6}$ & $\mathbf{93.8\pm1.9}$ & $\mathbf{86.7\pm1.4}$
& $\mathbf{85.9\pm3.8}$ & $\mathbf{73.8\pm2.6}$ & $\mathbf{87.4\pm1.7}$ \\
\bottomrule
\end{tabular}
}
\end{table*}

The best-checkpoint steps across the three seeds are GRPO 140/150/140, GiGPO 150/150/150, HiPER 90/145/145, and STAC 120/145/145.
The discrepancy between main and appendix summaries quantifies checkpoint-selection sensitivity rather than changing the method ranking.

\section{Boundary-Mask Representation}
\label{app:boundary-mask}

\paragraph{One turn.}
Let the response generated at environment turn $t$ contain $L_t$ tokens,
$y_t=(y_{t,1},\ldots,y_{t,L_t})$.
The parser locates the structured block that expresses the boundary decision, and $M_{t,j}^b=1$ exactly for token positions belonging to that block; every other position receives zero.
For example, suppose a tokenized response is schematically
\[
\underbrace{\langle\mathrm{switch}\rangle\ \mathrm{KEEP}\
\langle/\mathrm{switch}\rangle}_{\text{boundary block}}\quad
\underbrace{\langle\mathrm{subgoal}\rangle\ \cdots}_{\text{subgoal block}}\quad
\underbrace{\langle\mathrm{action}\rangle\ \cdots}_{\text{action block}} .
\]
If the boundary block occupies the first three token positions, its mask is
\[
M_t^b=[1,\ 1,\ 1,\ 0,\ 0,\ldots,0].
\]
The exact number of ones depends on tokenization and on whether delimiter tokens are included in the configured mask.
If the boundary block is missing or malformed, the parser selects no boundary tokens and
\[
M_t^b=[0,\ 0,\ldots,0],
\qquad
\lVert M_t^b\rVert_0=0,
\]
so that turn contributes no boundary-level term.

\paragraph{A complete task trace.}
The mask is reconstructed from the response at every environment turn, not only when the agent creates a new subgoal.
Thus a $T$-turn task produces $M_1^b,\ldots,M_T^b$.
After padding each response to a common width $L$, the masks form
\[
M^b\in\{0,1\}^{T\times L}.
\]
A schematic four-turn trace is
\[
M^b=
\begin{bmatrix}
1&1&1&0&0&\cdots\\
1&1&1&0&0&\cdots\\
1&1&1&0&0&\cdots\\
0&0&0&0&0&\cdots
\end{bmatrix}.
\]
The first three rows contain valid boundary blocks; each may encode either \keep{} or \switch{}.
The fourth response has no valid boundary block, so its row is all zero.
Consequently, $M_t^b$ records \emph{where the current turn writes its boundary decision}; it does not identify the active subgoal or persist for the lifetime of that subgoal.
Subgoal tokens are represented by a separate mask.


\end{document}